\documentclass[sigconf]{acmart}
\AtBeginDocument{%
  }

\usepackage{hyperref}
\usepackage{colortbl}

\usepackage{enumitem}

\usepackage{pifont}
\newcommand{\cmark}{\ding{51}}%
\newcommand{\xmark}{\ding{55}}%

\definecolor{lightgray}{gray}{0.9}

\copyrightyear{2026}
\acmYear{2026}
\setcopyright{cc}
\setcctype{by-nc-nd}
\acmConference[ICMR '26]{International Conference on Multimedia Retrieval}{June 16--19, 2026}{Amsterdam, Netherlands}
\acmBooktitle{International Conference on Multimedia Retrieval (ICMR '26), June 16--19, 2026, Amsterdam, Netherlands}

\acmISBN{979-8-4007-2617-0/2026/06}
\acmDOI{10.1145/3805622.3810590}     
\acmSubmissionID{7466}

\begin{document}

\title{VietFashion: Benchmarking Sketch–Text Composed Image Retrieval for Cultural Outfits}

% ---------------- AUTHORS ----------------
\author{Hoang-Nguyen Cao}
\orcid{0009-0003-0459-730X}
\authornote{Both authors contributed equally to this research.}
\affiliation{
  \institution{University of Science}
  \city{Ho Chi Minh City}
  \country{Vietnam}
}
\affiliation{
  \institution{Vietnam National University}
  \city{Ho Chi Minh City}
  \country{Vietnam}
}
\email{chnguyen23@apcs.fitus.edu.vn}

\author{Le-Hoang Bui}
\orcid{}
\authornotemark[1]
\affiliation{
  \institution{University of Science}
  \city{Ho Chi Minh City}
  \country{Vietnam}
}
\affiliation{
  \institution{Vietnam National University}
  \city{Ho Chi Minh City}
  \country{Vietnam}
}
\email{blhoang23@apcs.fitus.edu.vn}

\author{Dinh-Khoi Vo}
\orcid{0000-0001-8831-8846}
\affiliation{
  \institution{University of Science}
  \city{Ho Chi Minh City}
  \country{Vietnam}
}
\affiliation{
  \institution{Vietnam National University}
  \city{Ho Chi Minh City}
  \country{Vietnam}
}
\email{vdkhoi@selab.hcmus.edu.vn}

\author{Minh-Triet Tran}
\orcid{0000-0003-3046-3041}
\affiliation{
  \institution{University of Science}
  \city{Ho Chi Minh City}
  \country{Vietnam}
}
\affiliation{
  \institution{Vietnam National University}
  \city{Ho Chi Minh City}
  \country{Vietnam}
}
\email{tmtriet@fit.hcmus.edu.vn}

\author{Trung-Nghia Le}
\orcid{0000-0002-7363-2610}
\affiliation{
  \institution{University of Science}
  \city{Ho Chi Minh City}
  \country{Vietnam}
}
\affiliation{
  \institution{Vietnam National University}
  \city{Ho Chi Minh City}
  \country{Vietnam}
}
\email{ltnghia@fit.hcmus.edu.vn}
\authornote{Corresponding author.}

\renewcommand{\shortauthors}{Hoang-Nguyen Cao et al.}

% ---------------- CCS TERMS ----------------


% ---------------- KEYWORDS ----------------

% ---------------- ABSTRACT ----------------
\begin{abstract}
    Cultural garments pose a unique challenge for visual retrieval systems, as their identity often depends on subtle structural and symbolic details that are poorly captured by standard AI models. We introduce VietFashion, a new benchmark for sketch–text composed image retrieval centered on the Ao Dai, a traditional Vietnamese garment. VietFashion enables designers and researchers to retrieve culturally meaningful outfits using a combination of hand-drawn sketches, which convey garment structure, and textual descriptions, which encode cultural semantics. The dataset is initialized with 650 sketches and expanded using generative models to produce over 21,000 photorealistic images with aligned captions. Textual prompts that describe detailed outfit attributes, which are extracted from fashion magazines to ensure authenticity and diversity. To better reflect the inherent ambiguity of design intent, VietFashion adopts a multi-target retrieval setting, where a single query may correspond to multiple valid results. We establish standardized evaluation protocols and benchmark state-of-the-art composed image retrieval methods. Experimental results reveal significant performance gaps in modeling fine-grained cultural semantics and multi-modal composition, positioning VietFashion as a challenging benchmark for fine-grained fashion retrieval. The dataset is publicly available at: \url{https://hng0303.github.io/VietFashion}.
    
    % VietFashion is constructed using hand-drawn sketches paired with textual prompts describing detailed outfit attributes, including silhouette, collar type, fabric, patterns, and cultural motifs. These attributes are crawled from fashion magazines to ensure authenticity and diversity. Leveraging generative AI, we synthesize culturally consistent Ao Dai images conditioned on sketch–text inputs, enabling controlled variations across styles and contexts. The dataset supports multiple retrieval settings, including traditional image retrieval, compositional image retrieval (CIR), and zero-shot retrieval, with a particular emphasis on compositional sketch–text image retrieval, a challenging yet underexplored problem. We establish standardized evaluation protocols and benchmark representative traditional, CIR, and zero-shot methods. Experimental results reveal significant performance gaps in modeling fine-grained cultural semantics and multi-modal composition, positioning VietFashion as a challenging benchmark for future research.
\end{abstract}
% ---------------- FIGURE AFTER AUTHORS ----------------
\begin{teaserfigure}
\centering
\includegraphics[width=\textwidth]{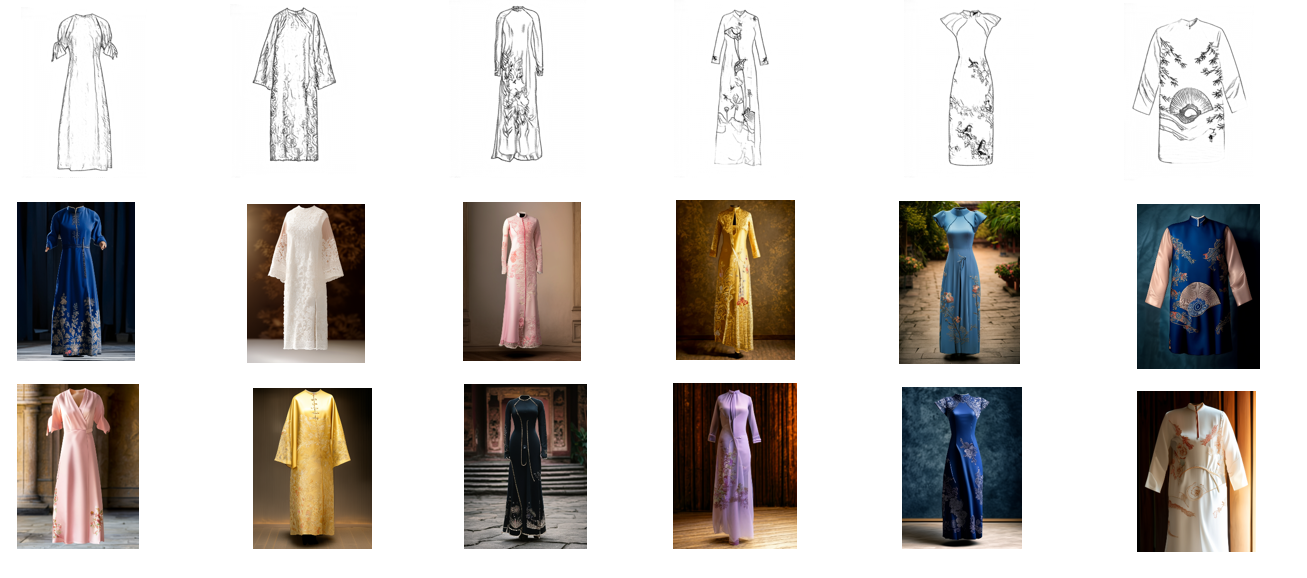}
\caption{
Representative sketch-photo pairs of Ao Dai from our VietFashion dataset. The top row features sketches reflecting diverse levels of abstraction and detail, while the bottom rows present corresponding photographs of authentic garments.
}
\label{fig:samples}
\end{teaserfigure}

\maketitle

% ---------------- MAIN CONTENT ----------------
\section{Introduction}

% \highlight{bo sung reference cho section nay} \highlight{KHONG duoc dung GPT de sinh reference, kiem tra cac reference CAN THAN} \highlight{Dung ZeroGPT de check cac reference}

The digitalization of cultural heritage is an urgent task, yet modern image retrieval systems often fall short in this area. While significant progress has been made in visual retrieval~ \cite{baldrati2023composed, saito2023pic2word, baldrati2023zero}, current systems predominantly focus on Western fashion or broad, generic categories like "dress" or "shirt"~ \cite{wu2021fashion, garderes2025facap}. As a result, they fail to capture the highly fine-grained details of traditional craftsmanship, such as specific collar cuts, complex embroidery motifs, and unique fabric drapes, that define a culture's visual identity. This technology gap hinders efforts to preserve these traditional designs and restricts creative designers who want to modernize or explore variations of cultural garments.

A particularly promising solution to this problem is Sketch–Text Compositional Image Retrieval (ST-CIR)~ \cite{sangkloy2022taskformer, gatti2024composite}. In this setting, a user provides a hand-drawn sketch to establish the structural bones of an outfit (e.g., the silhouette and layout) and uses a text prompt to add the cultural flesh (e.g., textures, colors, and specific motifs). This combination offers a powerful tool for design exploration, allowing users to search for specific cultural attributes that are difficult to describe with words alone or draw with perfect accuracy.

% Multi-modal image retrieval has made significant progress, yet existing research primarily focuses on photo-based or text-only queries. A particularly challenging and practically relevant setting is compositional sketch–text image retrieval (ST-CIR), where a user specifies visual intent through a hand-drawn sketch and refines it with textual attributes. While sketches provide the structural "bones" of a visual intent—capturing silhouette and spatial layout—textual descriptions provide the "flesh," detailing nuances like texture, color, and cultural motifs. By combining the structural intuition of a hand-drawn sketch with the descriptive precision of natural language, CSTBIR offers an unparalleled level of expressiveness, making it a highly potential frontier for personalized and professional search engines.

% Despite its relevance, the path toward effective Compositional Sketch-Text Image Retrieval (ST-CIR) remains obstructed by three critical research gaps. 

However, advancing ST-CIR for cultural heritage faces major barriers. There is a significant architecture mismatch in current research; most CIR models are optimized for image-text inputs and struggle to connect sparse, abstract sketches with dense, colorful photos~ \cite{baldrati2023composed, consensus_cir}. Unlike natural photos, sketches lack dense pixel information, making it difficult for standard vision-language models to extract meaningful features~ \cite{liu2017deep, sain2023clip}. Moreover, there is a lack of high-quality datasets that pair sketches with culturally grounded text attributes. Existing datasets like FashionIQ are predominantly Western-centric~ \cite{wu2021fashion}. Without a ground truth that connects a structural sketch to fine-grained textual modifiers, training models to navigate complex compositional queries becomes challenging~ \cite{liu2021image, baldrati2023zero}.

% Third, existing evaluation metrics are often too coarse, prioritizing general category matches (e.g., identifying a "dress") over fine-grained correctness. In culturally sensitive domains, a retrieval is only truly successful if it respects both the structural intent of the sketch and the specific cultural attributes, such as a unique collar style or traditional motif, defined in the text.

To address this gap, we present VietFashion, a benchmark designed to push the boundaries of fine-grained, culturally-grounded retrieval. We focus on the Ao Dai, the traditional Vietnamese garment~ \cite{hanoivoyage_aodai, vietdreamtravel_aodai}, which serves as an ideal subject due to its strict structural elegance and a vast library of subtle, attribute-based variations. We aim to bridge the gap between manual creativity and synthetic diversity. To ensure our dataset is culturally accurate, we curated a rich vocabulary of descriptors (e.g., "raglan sleeves," "lotus motifs") directly from fashion archives and magazines. This ensures that the text component serves as a linguistic archive of traditional design terms. Collecting thousands of sketch-photo pairs for traditional garments is difficult. Rather than relying on limited web-scraped data, we leverage generative models to synthesize high-fidelity target images. We combine sketches with Qwen-2.5 (for text) and SANA-ControlNet (for images) to synthesize 21,000 photorealistic target images~ \cite{yang2025qwen3, xie2024sana}. We create a multi-target environment where retrieval models must distinguish between highly similar images based on subtle textual cues (Fig.~\ref{fig:samples}). Unlike older datasets that force a single correct answer for a query, we adopt a multi-target query design similar to the CIRCO benchmark~ \cite{baldrati2023zero}. For every query, we generate three distinct valid target images. This reduces false-negative errors during training and better models the real-world design process, where one concept can lead to multiple valid interpretations.

To assess the dataset's utility for cultural preservation and design, we established a rigorous benchmark suite evaluating four distinct retrieval paradigms: supervised composed image retrieval (CIR)~ \cite{baldrati2023composed, saito2023pic2word}, zero-shot CIR (ZS-CIR)~ \cite{zs_sbir_stylegen, lin2023zero}, sketch-text CIR (ST-CIR)~ \cite{sangkloy2022taskformer}, and sketch-based image retrieval (SBIR)~ \cite{eitz2012tuberlin, song2016sketchy, dey2019doodle2search}. Our experiments reveal that even supervised methods still remain low in recall. This underscores the significant challenge current models face in distinguishing fine-grained cultural semantics, such as specific embroidery patterns or collar styles, from abstract sketches. By providing a dataset where every sketch is paired with precise, attribute-driven modifications, our VietFashion offers a new standard for evaluating how well AI can understand the interplay between the stroke of a pen and the specificity of a word.

% We evaluate ST-CIR through three lenses: Traditional retrieval (feature-based), state-of-the-art CIR methods, and Zero-shot capabilities of large vision-language models.

% 

Our contributions are as follows:
\begin{itemize}
    \item We introduce VietFashion, a benchmark designed for cultural preservation that pairs sketches with magazine-grounded text attributes to capture the fine-grained semantics of the Vietnamese Ao Dai.
    
    \item We develop a generative pipeline utilizing Qwen-2.5 and SANA-ControlNet to synthesize fashion images, bridging the gap between manual creativity and data scale.
    
    \item We implement a multi-target query design that provides three valid ground-truth images per sketch, reducing the triplet ambiguity problem found in previous single-target benchmarks by modeling realistic design variations. We also establish a rigorous benchmark for multi-target retrieval through various retrieval paradigms.
    
    \item The VietFashion dataset is publicly available at: \url{https://hng0303.github.io/VietFashion}.
    
\end{itemize}

\section{Related Work}

\begin{table*}[t!]
\centering
\caption{Comparison of VietFashion with existing image retrieval dataset. VietFashion uniquely targets cultural outfits and employs multi-target supervision to address the ambiguity inherent in fine-grained sketch-text retrieval. 
% \highlight{kiem tra CAN THAN tung phuong phap xem thong tin co dung hay khong}
}
\label{tab:datasets}
\begin{tabular}{lcllcc}
\toprule
\multicolumn{1}{c}{\textbf{Dataset}} & \multicolumn{1}{c}{\textbf{Year}} & \multicolumn{1}{c}{\textbf{Domain}} & \multicolumn{1}{c}{\textbf{Query Modality}} & \multicolumn{1}{c}{\textbf{CIR}} & \multicolumn{1}{c}{\textbf{Multi-Target Retrieval}} \\
\midrule

TU-Berlin~ \cite{eitz2012tuberlin} & 2012 & General Object & Sketch & \xmark & \xmark \\

Sketchy Extended~ \cite{song2016sketchy} & 2016 & General Object & Sketch & \xmark & \xmark \\

FashionIQ~ \cite{wu2021fashion} & 2019 & Western Fashion & Image, Text & \cmark & \xmark \\

QuickDraw-Ext~ \cite{dey2019doodle2search} & 2019 & General Object & Sketch & \xmark & \xmark \\

CIRR~ \cite{liu2021image} & 2021 & Open-Domain & Image, Text & \cmark & \xmark \\

CIRCO~ \cite{baldrati2023zero} & 2022 & Open-Domain & Image, Text & \cmark & \cmark \\

FACap~ \cite{garderes2025facap} & 2025 & Fashion & Image, Text & \cmark & \cmark \\

CSTBIR~ \cite{gatti2024composite} & 2025 & General Object & Sketch, Text & \cmark & \xmark \\

FIGROTD~ \cite{le2026figrotd} & 2026 & General Object & Sketch, Text, Image & \cmark & \xmark \\

\rowcolor{lightgray} VietFashion (Ours) & 2026 & Cultural Outfit & Sketch, Text & \cmark & \cmark \\
\bottomrule
\end{tabular}
\end{table*}

\subsection{Existing Image Retrieval and Cultural Fashion Datasets}

Table~\ref{tab:datasets} summarizes the landscape of existing benchmarks supporting sketch, composed, and multimodal retrieval. In the domain of sketch-based image retrieval (SBIR), foundational datasets range from TU-Berlin~ \cite{eitz2012tuberlin} to larger collections like Sketchy Extended~ \cite{song2016sketchy} and QuickDraw-Extended~ \cite{dey2019doodle2search}. For composed image retrieval (CIR), FashionIQ~ \cite{wu2021fashion} and CIRR~ \cite{liu2021image} established early standards, while recent works such as CIRCO~ \cite{baldrati2023zero} and FACap~ \cite{garderes2025facap} have introduced multi-target supervision and large-scale automated generation to address retrieval ambiguity and data scarcity. Bridging these modalities, FIGROTD~ \cite{le2026figrotd} supports unified sketch, text, and composed queries, while CSTBIR~ \cite{gatti2024composite} offers a massive scale of multimodal queries. Collectively, these datasets provide essential benchmarks for evaluating diverse retrieval paradigms, setting the stage for our focused contribution to cultural heritage.

% Several datasets support sketch, composed, and multimodal retrieval research.

% For SBIR, TU-Berlin~ \cite{eitz2012tuberlin} contains 20K sketches across 250 categories. Sketchy Extended~ \cite{sketchy_extended} contains 75K sketches and 73K photos across 125 categories. QuickDraw-Extended~ \cite{dey2019doodle2search} contains 330K sketches and 204K photos across 110 categories.

% For composed retrieval, FashionIQ~ \cite{wu2021fashion} contains approximately 18K composed query triplets. CIRR~ \cite{liu2021image} provides composed retrieval for open-domain objects. CIRCO~ \cite{baldrati2023zero} introduces multi-target supervision for composed retrieval evaluation. FACap~ \cite{garderes2025facap} provides over 227K automatically generated fashion CIR triplets. FIGROTD~ \cite{le2026figrotd} provides approximately 16.5K multimodal retrieval triplets supporting sketch, text, and composed query modes.

% For compositional sketch-text retrieval, CSTBIR~ \cite{gatti2024composite} provides approximately 2M multimodal queries paired with 108K images, representing one of the largest multimodal retrieval datasets currently available.

% These datasets provide essential benchmarks for evaluating multimodal retrieval models across sketch, text, and composed retrieval paradigms.

% \noindent\textbf{Motivation and Our Contribution.}

Despite the progress in multimodal retrieval, existing datasets exhibit several limitations when applied to culturally specific fashion domains. First, most fashion CIR and SBIR datasets focus on Western clothing styles and lack representation of traditional garments. Second, many benchmarks assume single-target retrieval, which creates ambiguity during training because visually similar garments may all be correct retrieval results. Third, existing datasets rarely combine sketch and natural-language queries for traditional fashion items, limiting the evaluation of multimodal retrieval models in real-world use cases.

In parallel, several works have explored datasets centered on culturally grounded garments and heritage fashion. For example, HanfuBench~ \cite{zhou2025hanfu} focuses on traditional Chinese Hanfu garments and supports tasks such as cultural understanding and visual reasoning. Similarly, EUFCC-CIR~ \cite{net2024eufcc} introduces a composed image retrieval benchmark for cultural heritage collections derived from museum archives, while WCCA-AK~ \cite{oh2025wcca} provides a multimodal dataset for haute couture preservation and analysis. These datasets highlight the growing importance of cultural representation in vision-language research. However, they are primarily designed for tasks such as classification, visual understanding, or generative modeling, and do not explicitly address fine-grained retrieval under sketch–text compositional settings.

In contrast, VietFashion is specifically designed as a benchmark for sketch–text composed image retrieval in a culturally grounded domain. It integrates (i) sketches as structural queries, (ii) text descriptions capturing fine-grained cultural attributes, and (iii) a multi-target retrieval protocol that models real-world ambiguity. This combination distinguishes VietFashion from both general retrieval benchmarks and existing cultural heritage datasets, providing a unified framework to study fine-grained cultural garment retrieval.

% To address these limitations, we introduce a new sketch-text retrieval dataset for Vietnamese traditional clothing (\textit{\'a}o d\`ai). The dataset contains 650 sketches paired with natural-language captions and three corresponding target fashion images per query. Through controlled synthesis and data augmentation, we construct approximately 21,000 synthesized fashion images and 21,000 composed retrieval triplets. Unlike conventional single-target retrieval datasets, our dataset adopts a multi-target query design that explicitly mitigates the triplet ambiguity problem by allowing multiple valid retrieval results.

% We further develop an automated data generation pipeline combining large language models and vision generation models. Captions are generated using Qwen-2.5, while target images are synthesized using the SANA vision model conditioned on garment attributes. Finally, we benchmark multiple baseline and state-of-the-art retrieval models using standard metrics including Recall@K, mAP, and MRR under sketch–text composed retrieval settings.

\subsection{Composed Image Retrieval (CIR)}

CIR aims to retrieve a target image given a query composed of a reference image and a textual modification, allowing users to express fine-grained search intent such as attribute editing or style transformation. Early CIR benchmarks such as FashionIQ~ \cite{wu2021fashion} and CIRR~ \cite{liu2021image} established standard evaluation protocols for composed retrieval across fashion and general object domains. FashionIQ contains approximately 18K triplets across clothing categories and supports relative attribute-based retrieval, while CIRR extends composed retrieval to open-domain objects via natural-language modification queries.

Despite their importance, early CIR datasets are limited in scale and semantic diversity, which restricts model generalization and compositional reasoning. To address these limitations, recent research focuses on constructing large-scale datasets and developing automatic annotation pipelines. The FACap dataset~ \cite{garderes2025facap} introduces more than 227K automatically constructed fashion CIR triplets using web-scale image crawling combined with vision-language model filtering, significantly expanding training diversity and enabling large model training. Similarly, the FIGROTD benchmark~ \cite{le2026figrotd} introduces an image-guided retrieval task setting supporting sketch-only, text-only, and composed queries, with approximately 16.5K training triplets.

Another important development is multi-target CIR evaluation. The CIRCO dataset~ \cite{baldrati2023zero} introduces multiple valid target images per query, reducing false-negative supervision and improving evaluation robustness. ZS-CIR frameworks further reduce annotation cost by mapping reference image features into shared token or embedding spaces and combining them with textual modifications to perform retrieval without explicit triplet annotation~ \cite{baldrati2023zero}.
In fashion-specific retrieval, multimodal alignment remains challenging due to fine-grained garment attributes such as fabric, texture, and structure. Recent works leverage vision-language foundation models and graph-based alignment frameworks to improve cross-modal retrieval performance. For example, graph-integrated cross-modal learning improves structural semantic reasoning in fashion retrieval~ \cite{graph_fashion_retrieval_2025}, while self-supervised CLIP-based approaches enable scalable cross-domain representation learning~ \cite{sfclip_2024}. Multi-hop reasoning models further improve compositional fashion retrieval by modeling attribute relationships across semantic reasoning steps~ \cite{cfir_multihop_2025}.

\subsection{Sketch-Based Image Retrieval (SBIR)}

SBIR focuses on retrieving natural images using free-hand sketches. The main challenge lies in the large domain gap between abstract human sketches and realistic natural images. Early SBIR benchmarks such as TU-Berlin~ \cite{eitz2012tuberlin} and Sketchy~ \cite{song2016sketchy} enabled large-scale category-level sketch-photo retrieval evaluation. TU-Berlin contains approximately 20,000 sketches across 250 object categories, while Sketchy Extended contains over 75K sketches and 73K photos across 125 categories~ \cite{song2016sketchy}.

Early SBIR research primarily focused on category-level retrieval using hand-crafted descriptors and bag-of-features representations. Deep learning-based SBIR later shifted toward shared-embedding learning and domain-adaptation techniques. Deep Sketch Hashing introduced fast retrieval using learned compact binary embeddings~ \cite{liu2017deep}.

Recent SBIR research focuses on zero-shot and large-scale retrieval. The Doodle2Search framework~ \cite{dey2019doodle2search} introduces the QuickDraw-Extended dataset, which contains approximately 330K sketches and 204K photos across 110 categories, and proposes a zero-shot retrieval framework based on semantic embedding alignment. Zero-shot SBIR frameworks using semantic adversarial learning further improve retrieval robustness by aligning sketch features with semantic word embeddings~ \cite{xu2019semantic}.

Zero-shot SBIR approaches extend generalization capability to unseen categories using semantic transfer and domain generalization techniques~ \cite{zs_sbir_stylegen}. Recent transformer-based SBIR approaches such as S3BIR-DINO~\cite{s3bir_dino_2025} feature adaptation further improve cross-domain representation robustness using self-supervised vision transformer features.

Despite these advances, SBIR remains challenging for fashion applications due to fine-grained garment structures, layered textures, and subtle attribute variations that sketches alone cannot fully represent.

\begin{figure*}
\centering
\includegraphics[width=\textwidth]{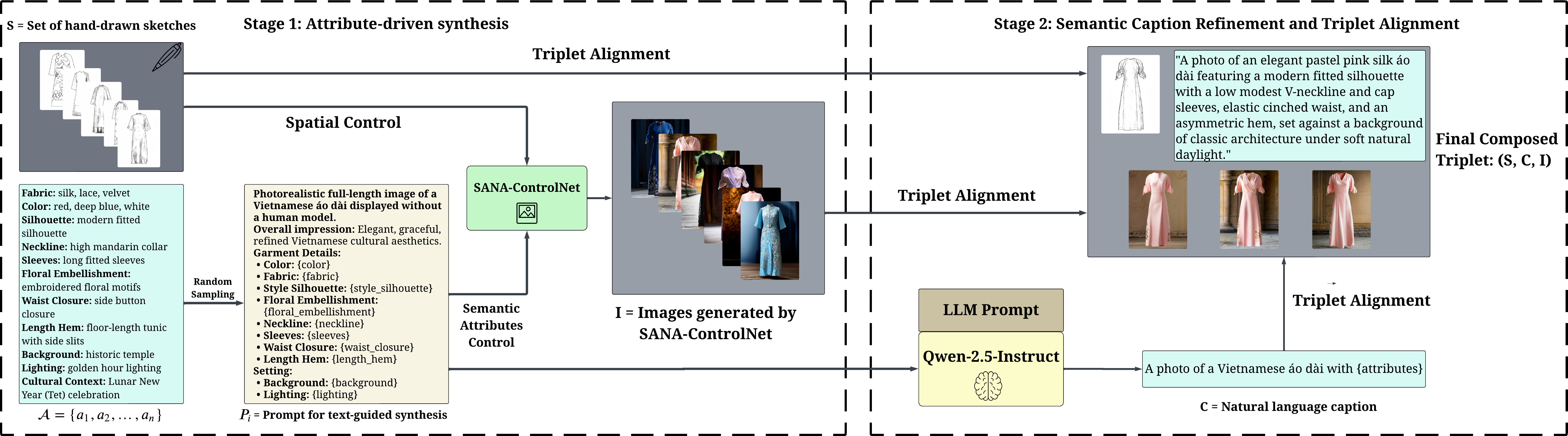}
\caption{
Overview of the VietFashion dataset construction pipeline. The pipeline begins with sketches (S) and sampled garment attributes (A). We utilize SANA-ControlNet to generate multi-target images (I) under spatial constraints, while Qwen-2.5-Instruct distills attributes into concise natural language captions (C) to form the final composed retrieval triplet.
% \highlight{chu qua nho. tang font size cua nhung phan quan trong. su dung PDF, khong xai PNG}
}
\label{fig:pipeline}
\end{figure*}

\subsection{Sketch--Text Composed Image Retrieval (ST-CIR)}

To overcome limitations of single-modality queries, recent research integrates sketch and text modalities to provide complementary structural and semantic information. TASK-former~ \cite{sangkloy2022taskformer} introduces a transformer-based dual encoder that jointly embeds sketch and text modalities and demonstrates significant retrieval improvement compared to text-only retrieval.

Large-scale multimodal datasets have been introduced to support compositional retrieval learning. CSTBIR~ \cite{gatti2024composite} provides approximately 2M sketch-text queries paired with 108K images, enabling large-scale multimodal training. The STNET multimodal transformer framework~ \cite{gatti2024composite} further improves retrieval performance by using sketch modality for spatial localization and text modality for attribute-level reasoning.

CLIP-based multimodal fusion frameworks align sketch, text, and image modalities into unified embedding spaces, enabling strong zero-shot generalization performance in multimodal retrieval tasks~ \cite{sain2023clip}. The FIGROTD benchmark further integrates SBIR and composed retrieval tasks into a unified evaluation framework supporting multiple query modalities~ \cite{le2026figrotd}.

Existing works demonstrate that sketch and text modalities provide complementary retrieval signals, where sketches capture spatial structure and text provides semantic attributes such as color, material, and contextual description. Building on this, we structure our benchmark to evaluate how effectively different methods leverage these distinct contexts. These comparisons enable us to quantify the specific value of textual attributes in bridging the gap between abstract sketches and culturally rich target images.

\section{VietFashion Dataset}

\subsection{Dataset Construction}

% \highlight{chua sua}

The proposed dataset is designed to support sketch--text composed fashion image retrieval in a culturally specific domain, focusing on the Vietnamese traditional garment (e.g., Ao Dai, or \'ao d\`ai). The dataset is constructed through a multi-stage pipeline combining human sketch collection, large language model (LLM) based caption generation, and vision-based image synthesis.

\textbf{Curated Attribute Collection}
To ensure cultural authenticity and linguistic diversity, we curated a vocabulary of garment attributes across eleven categories. These descriptors were systematically gathered through a process of targeted web crawling and manual curation from reputable fashion magazines and cultural archives, including Hanoi Voyage\cite{hanoivoyage_aodai}, Private Tour Asia\cite{privatetourasia_aodai}, and Viet Dream Travel\cite{vietdreamtravel_aodai}. By sourcing terminology directly from specialized fashion literature, the dataset achieves high cultural believability, reflecting real-world professional descriptors for the Vietnamese Ao Dai.

The attributes are organized into a hierarchical structure to facilitate fine-grained retrieval challenges:

\begin{itemize}
    \item Structural attributes: silhouette, neckline, and sleeves define the core spatial constraints.
    \item Detail attributes: fabric, floral embellishment, and waist and closure provide the semantic modifiers essential for distinguishing highly similar garments.
    \item Contextual attributes: background, lighting, and cultural context ground the synthesized images in realistic settings.
\end{itemize}

\textbf{Sketch Collection.}
We collected 650 sketches representing diverse Ao Dai designs. The sketches cover variations in garment silhouette, sleeve structure, collar type, decorative layout, and overall structural composition. This sketch set forms the structural query modality and simulates realistic user-generated sketch queries.

To improve structural diversity, sketches were curated to cover combinations of garment attributes including traditional and modern silhouettes, sleeve variants, and layered structural styles. This helps reduce structural bias and improves generalization ability during retrieval training.

\begin{table}[t!]
\centering
\caption{Example sketch metadata annotations illustrating variation in abstraction, structural complexity, and garment shape.}
\label{tab:sketch_metadata}
\begin{tabular}{lcccc}
\toprule
\textbf{Sketch ID} & \textbf{Abstraction} & \textbf{Structure} & \textbf{Width} & \textbf{Silhouette} \\
\midrule
0000001 & Low & High & Wide & Broad \\
0000006 & High & Medium & Wide & Broad \\
0000010 & Medium & Low & Regular & Balanced \\
\bottomrule
\end{tabular}
\end{table}

\paragraph{Sketch Metadata and Diversity.}
To further characterize the sketch modality, we analyze drawing composition, abstraction level, and structural complexity across all 650 sketches. The sketches follow a consistent composition (single centered garment on a plain background), reducing irrelevant variation and focusing on garment structure.

At the same time, abstraction and structural complexity vary meaningfully. The dataset is approximately balanced across abstraction levels (215 low, 214 medium, 221 high) and similarly spans simple to complex structural patterns (see Table~\ref{tab:sketch_metadata}). These properties are estimated using lightweight image-based measurements such as stroke density and connected components. This diversity improves robustness by requiring models to align both coarse silhouettes and fine-grained attributes across sketches of varying detail levels, supporting better generalization in real-world sketch-based queries.

\textbf{Caption Generation.}
To translate structural and stylistic requirements into natural language, we utilize a structured generation pipeline powered by the Qwen-2.5 3B Instruct model~ \cite{yang2025qwen3}. The process begins with the stochastic assembly of garment profiles; for each entry, we perform a random selection from our curated attribute lists, including fabric type (e.g., velvet, organza, lace), neckline (e.g., U-shaped, boat, rounded), sleeves (e.g., raglan, flared, cap), and silhouette. These attributes are then formatted into a comprehensive structured prompt that specifies the garment's appearance, material, and cultural context while explicitly excluding human features, mannequins, or photography-specific terminology.

The LLM is then tasked with condensing this structured input into a neutral, factual, and concise single-sentence caption starting with the phrase "A photo of". This approach ensures high semantic richness and consistent linguistic structure, resulting in an average caption length of approximately 42.46 words. 

\begin{table}[H]
\begin{tabular}{p{0.9\linewidth}}
\toprule
\textbf{Task: }You are given a structured fashion description of a Vietnamese áo dài (garment-only, no human features). Your task is to generate a short, concise image caption suitable for an image-text retrieval dataset starting with "A photo of". \\
\textbf{Rules (Strict): }Output exactly one sentence starting with "A photo of" Keep the caption neutral, factual, and descriptive, focusing only on garment visual attributes such as appearance, material, color, silhouette, and pattern. Include cultural context only if explicitly provided. Do not mention mannequins, displays, lighting, or photography terms. Avoid emotional or narrative wording and use clear, standard fashion vocabulary. \\
\textbf{Input:} \{prompt\} \\
\textbf{Output:} A short caption describing the Ao Dai as seen in the image. \\
\bottomrule
\end{tabular}
\end{table}

% \begin{figure}[H]
% \centering
% \includegraphics[width=\linewidth]{figures/prompts.png}
% % \caption{LLM-based caption generation component used in the dataset construction pipeline, showing the structured Ao Dai description input, rule-constrained prompt template, and the generated retrieval-ready caption output.}
% \label{fig:prompt_for_llm}
% \end{figure}

By leveraging this structured prompt-to-caption workflow, we produce a dataset of 7,000 composed retrieval queries that maintain a high degree of cross-modal alignment between the abstract sketch structure and detailed textual semantic attributes.

\textbf{Image Generation.}
Target fashion images are synthesized using the SANA-ControlNet model~ \cite{xie2024sana}. For each sketch-caption pair, three visually distinct but semantically consistent garment images are generated. These images vary in texture rendering, lighting, and minor visual details while preserving the core garment identity. This multi-instance generation improves retrieval robustness and reduces overfitting to single visual realizations. 

\begin{figure*}[t!]
\centering
\includegraphics[width=\textwidth]{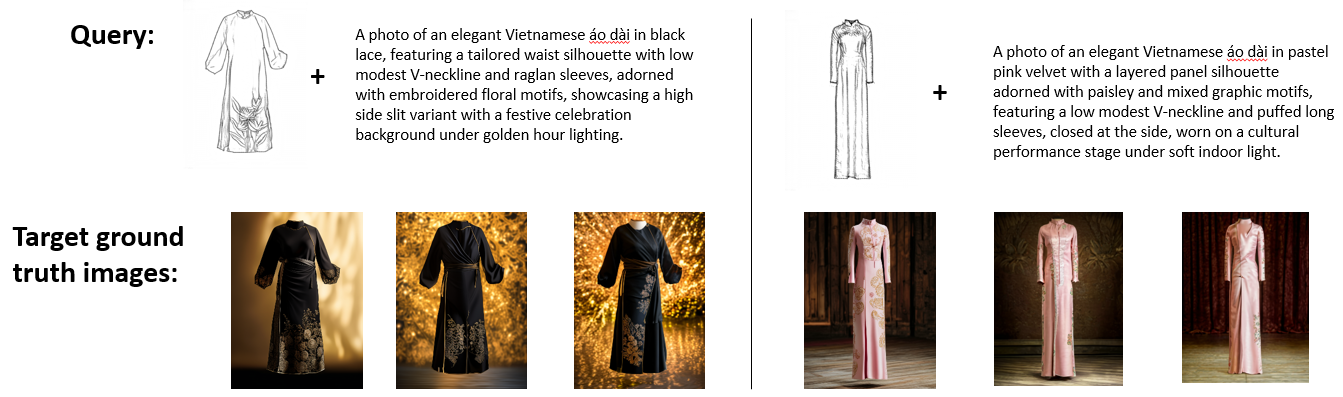}
\caption{Examples from the proposed VietFashion dataset. Each query contains a sketch of an Ao Dai, a natural-language caption describing garment attributes and context, and multiple valid target images.}
\label{fig:dataset_example}
\end{figure*}

\begin{figure}[t]
\centering
\includegraphics[width=\linewidth]{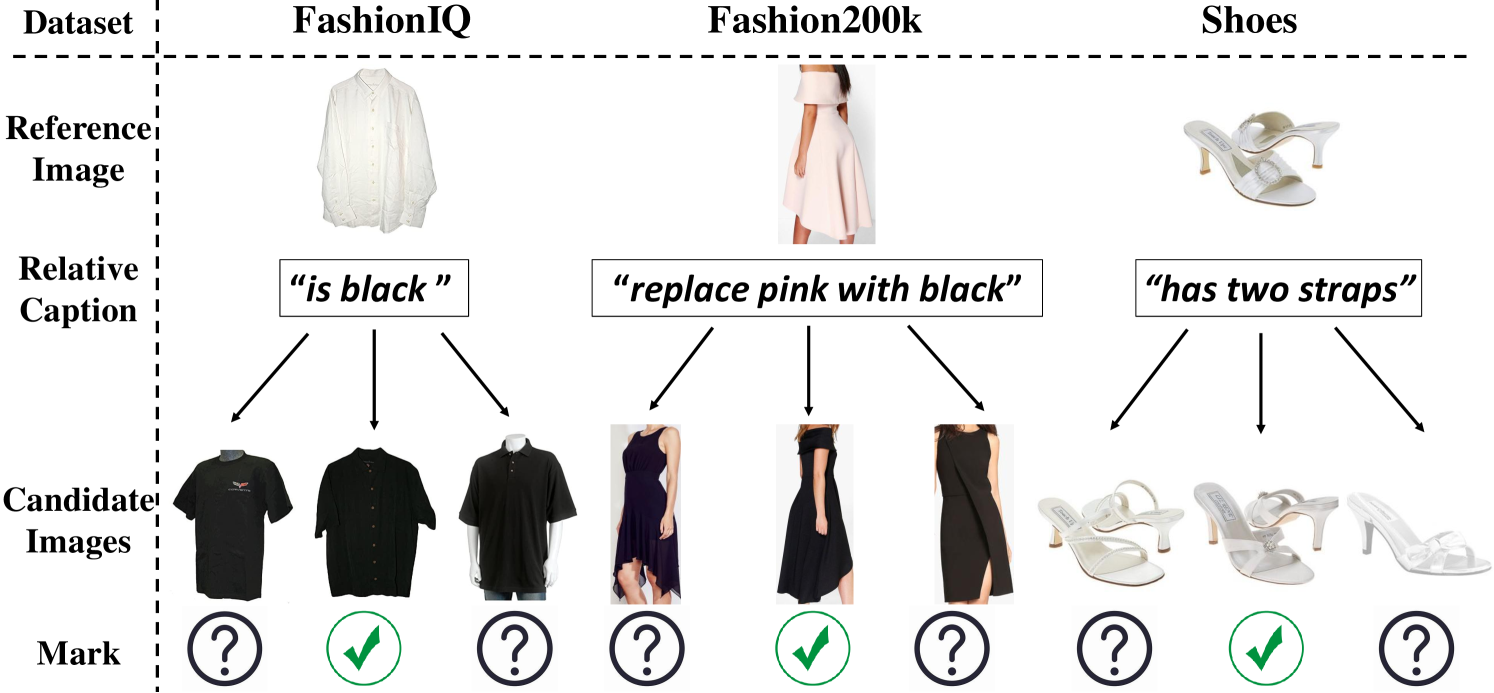}
\caption{Triplet ambiguity in CIR. A single query may correspond to multiple valid target images in existing fashion retrieval datasets. This figure is adapted from prior CIR literature addressing noisy supervision and consensus learning~ \cite{consensus_cir}.
}
\label{fig:triplet_ambiguity_example}
\end{figure}

\textbf{Multi-Target Query Design.} In CIR, a query is typically formed by combining a reference visual signal and an attribute modification. However, in real-world fashion scenarios, multiple target images may satisfy the same semantic description. When only one positive target is used during training, other valid targets may be misclassified as negatives, leading to false-negative supervision as in FashionIQ~\cite{wu2021fashion} and CIRR~\cite{liu2021image} datasets (see Fig.~\ref{fig:triplet_ambiguity_example}).

Recent works have highlighted this limitation and introduced datasets with multiple ground-truth targets per query to better model retrieval uncertainty and reduce training noise~ \cite{baldrati2023zero, garderes2025facap}. For example, the CIRCO dataset~\cite{baldrati2023zero} provides multiple valid targets for each query, explicitly addressing false-negative supervision in composed retrieval benchmarks. This reflects a broader shift toward modeling retrieval as a one-to-many mapping rather than one-to-one matching.

Our dataset follows a similar philosophy but applies it to a culturally specific fashion domain with compositional sketch--text queries. This design models realistic retrieval scenarios in which multiple visually similar garments may yield correct retrieval results. Each query is defined as:

$
(\text{Sketch}, \text{Caption}) \rightarrow \{\text{Target Img.}_1, \text{Target Img.}_2, \text{Target Img.}_3\}
$. 

\subsection{Dataset Description}

% \highlight{chua sua}

The proposed VietFashion dataset is specifically designed for ST-CIR of Vietnamese traditional garments. Each data sample consists of three tightly coupled modalities: structural sketch input, semantic textual description, and multiple valid visual targets. The dataset is organized into three main components, including 650 human sketches, 7,000 composed retrieval queries, and 21,000 synthesized fashion images. To prevent query-level leakage, the data is split at the sketch level into subsets, including 5,200 training queries, 650 validation queries, and 1,150 testing queries. Figure~\ref{fig:dataset_example} shows sample queries from our dataset. 

% Compared with existing fashion retrieval datasets, our dataset emphasizes three key aspects. First, it focuses on culturally specific garment structures. Most large-scale fashion retrieval datasets primarily contain Western clothing categories. In contrast, our dataset models traditional Vietnamese garment structure, decorative layout, and cultural usage context. Second, the proposed dataset supports compositional multimodal queries that combine sketch structure and textual attribute descriptions. This setting better simulates real user search behavior compared to image-only or text-only retrieval. This formulation is related to recent sketch-text composed retrieval frameworks~ \cite{sangkloy2022taskformer, gatti2024composite}. Third, our dataset adopts a multi-target supervision strategy. Each query is associated with three semantically consistent target images. These targets differ in rendering details, such as lighting, pose, and texture, while preserving the core garment identity.
VietFashion distinguishes itself from general fashion benchmarks through three core pillars. First, it prioritizes cultural-specific task formulation. While existing large-scale datasets predominantly feature Western clothing categories , VietFashion explicitly models the intricate structural elegance, specialized decorative layouts, and cultural usage contexts unique to the Vietnamese Ao Dai. Second, the dataset supports compositional multimodal queries that merge sketch structure with detailed textual attributes. This formulation more accurately mirrors real-world design and search behaviors compared to single-modality retrieval, aligning with recent ST-CIR frameworks~ \cite{sangkloy2022taskformer, gatti2024composite}. Finally, we utilize a multi-target supervision strategy. Each query is linked to three semantically consistent images that vary in rendering details, such as lighting, pose, and fabric texture, while strictly preserving the garment's underlying cultural identity.

\section{Benchmark Suite}

% \highlight{chua sua}

To evaluate the effectiveness of various retrieval paradigms on the VietFashion dataset, we establish a comprehensive benchmark suite. This suite encompasses standard retrieval metrics and a diverse set of baseline methods, ranging from supervised compositional models to zero-shot approaches.
\subsection{Evaluation Metrics}
Following the protocol established in~ \cite{Tran_2025}, we use widely recognized metrics to assess the ranking quality and accuracy of retrieval systems:

\paragraph{Recall@K (R@1, R@5, R@10)} This metric measures the percentage of queries for which the correct target image is retrieved within the top $K$ results. $R@1$ represents the most stringent accuracy measure, while $R@5$ and $R@10$ provide insight into the model's ability to rank the target within a reasonable search window.

\paragraph{Mean Average Precision (mAP)} Unlike Recall, which treats all ranks within $K$ equally, $mAP$ accounts for the entire precision-recall curve. It provides a comprehensive score that rewards models for placing relevant images at the highest possible ranks.

\paragraph{Mean Reciprocal Rank (MRR)} $MRR$ focuses on the rank of the first relevant item. It is the average of the reciprocal ranks ($1/\text{rank}$) for all queries. A high $MRR$ indicates that the system consistently places the correct target near the top of the results, which is crucial for user satisfaction in real-world search workflows.

\subsection{Baseline Methods}

To comprehensively evaluate the VietFashion benchmark, we consider multiple retrieval paradigms spanning sketch-only retrieval, sketch-text composed retrieval, zero-shot adaptation, and supervised composed retrieval. This paper provides insight into how different levels of supervision and modality fusion influence retrieval performance in a culturally specific garment domain. We evaluate a diverse set of baselines on the VietFashion dataset under varying degrees of supervision and modality fusion. This evaluation enables a systematic analysis of:
\begin{itemize}
    \item Sketch-only structural retrieval capability
    \item Cross-modal compositional reasoning performance
    \item Zero-shot adaptation ability of foundation models
    \item Fully supervised upper-bound retrieval performance
\end{itemize}

\subsubsection{Sketch-Based Image Retrieval (SBIR)}

These methods use only the sketch modality as input and retrieve images based on structural similarity. They serve as lower bounds for quantifying the contribution of textual attribute guidance.

\begin{itemize}
    \item \textbf{ZSE-SBIR~ \cite{lin2023zero}:} A zero-shot sketch-based retrieval framework designed to generalize to unseen visual categories by aligning sketch and photo embeddings using semantic supervision.
    
    \item \textbf{S3BIR-DINO~ \cite{s3bir_dino_2025}:} A self-supervised SBIR approach that leverages DINO visual representations to improve sketch-photo feature alignment without requiring paired supervision.
\end{itemize}

\subsubsection{Sketch-Text Composed Image Retrieval (ST-CIR)}

These models explicitly fuse sketch structure and textual attribute descriptions to perform composed retrieval.

\begin{itemize}
    \item \textbf{TaskFormer~ \cite{sangkloy2022taskformer}:} A transformer-based architecture designed for joint reasoning over sketch and textual modalities. It learns compositional representations that encode both structural sketch information and semantic attribute modifications.

    \item \textbf{VaGFeM~ \cite{le2026figrotd}:} A recent compositional retrieval framework that introduces visual-guided feature modulation to improve cross-modal alignment between reference sketches and textual attributes. The model emphasizes fine-grained attribute control and cross-modal consistency.
\end{itemize}

\subsubsection{Supervised Composed Image Retrieval (CIR)}

These methods are trained using explicit triplet supervision (i.e., sketch, text, target image) and represent the strongest supervised baselines.

\begin{itemize}
    \item \textbf{CLIP4CIR~ \cite{baldrati2023composed}:} A framework uses CLIP as the backbone and employs a feature combiner module to fuse sketch and text embeddings into a unified retrieval representation.

    \item \textbf{BLIP4CIR~ \cite{li2022blip}:} A framework is built upon BLIP multimodal encoders, which provide stronger fine-grained text-image alignment, and integrates a combiner module for compositional retrieval.
\end{itemize}

\subsubsection{Zero-Shot and Textual Inversion-Based Methods}

These approaches evaluate the capability of general multimodal foundation models to perform composed retrieval without explicit triplet supervision.

\begin{itemize}
    \item \textbf{SEARLE-ViT/B~ \cite{baldrati2023zero}:} A textual inversion approach for zero-shot composed retrieval. It maps visual features of the reference sketch into pseudo-word tokens within the CLIP embedding space and combines them with textual attributes during retrieval.

    \item \textbf{SEARLE-ViT/L~ \cite{baldrati2023zero}:} A larger backbone variant providing stronger semantic alignment due to higher representation capacity.

    \item \textbf{Pic2Word~ \cite{saito2023pic2word}:} A framework converts reference sketches into pseudo-word embeddings that can be appended to text queries for standard text-to-image retrieval. We also employ a domain-adapted version where the mapping network is fine-tuned on our training set to better capture domain-specific structure and cultural garment patterns.
\end{itemize}

\begin{table*}[t!]
\centering
\caption{Retrieval performance on the VietFashion benchmark. Methods are categorized by learning paradigm. Higher values indicate better performance. Pic2Word (Fine-tuned) was adapted using sketches in our training set.}
\label{tab:results}
\begin{tabular}{@{}llccccc@{}}
\toprule
\textbf{Method} & \textbf{Paradigm} & \textbf{R@1} & \textbf{R@5} & \textbf{R@10} & \textbf{mAP} & \textbf{MRR} \\ \midrule
% \textit{Sketch-only (SBIR)} & & & & & & \\
ZSE-SBIR~\cite{lin2023zero} & SBIR & 0.0285 & 0.0623 & 0.1077 & 0.0323 & 0.0539 \\
S3BIR-DINO~\cite{s3bir_dino_2025} & SBIR & 0.0157 & 0.0565 & 0.0948 & 0.0216 & 0.0428 \\ 
% \midrule
% \textit{Sketch-text CIR} & & & & & & \\
Taskformer~\cite{sangkloy2022taskformer} & ST-CIR & 0.0564 & 0.1472 & 0.2067 & 0.0269 & 0.0891 \\
VaGFeM~\cite{le2026figrotd} & ST-CIR & 0.0750  & 0.1612 & 0.2201 & 0.0356 & 0.1142 \\ 
\midrule
% \textit{Supervised CIR} & & & & & & \\
CLIP4CIR~\cite{baldrati2023composed} & Supervised & 0.0313 & 0.1149 & 0.1851 & 0.1064 & 0.1908 \\
BLIP4CIR~\cite{li2022blip} & Supervised & 0.0877 & 0.2672 & 0.3703 & 0.2483 & 0.3950 \\ 
% \midrule
% \textit{Zero-shot / Inversion} & & & & & & \\
SEARLE-ViT/B~\cite{baldrati2023zero} & Zero-shot & 0.0000 & 0.0200 & 0.0400 & 0.0200 & 0.0500 \\
SEARLE-ViT/L~\cite{baldrati2023zero} & Zero-shot & 0.0100 & 0.0300 & 0.0400 & 0.0300 & 0.0600 \\
Pic2Word~\cite{saito2023pic2word} & Zero-shot & 0.0082 & 0.0210 & 0.0364 & 0.0253 & 0.0523 \\
Pic2Word~\cite{saito2023pic2word} (Fine-tuned) & Fine-tuned & 0.0087 &	0.0221 & 0.0374	& 0.0253 & 0.0527\\ \bottomrule
\end{tabular}
% \vspace{2pt}
% \begin{flushleft}
% % \small{*Note: }
% \end{flushleft}
\end{table*}

\subsection{Quantitative Results}

The quantitative performance of the evaluated baseline models on the VietFashion benchmark is summarized in Table~\ref{tab:results}. 

Overall, supervised CIR methods achieve the strongest performance. BLIP4CIR~ \cite{li2022blip} establishes the best baseline across all evaluation metrics, achieving Recall@1 of 0.0877, Recall@10 of 0.3703, mAP of 0.2483, and MRR of 0.3950. CLIP4CIR~ \cite{baldrati2023composed} also performs competitively but remains significantly below BLIP4CIR, particularly in ranking-based metrics such as MRR (0.1908 vs 0.3950).

Meanwhile, zero-shot and inversion-based methods exhibit the lowest performance. SEARLE-ViT/L~ \cite{baldrati2023zero} provides the strongest zero-shot baseline with Recall@1 of 0.0100 and MRR of 0.0600. Pic2Word~ \cite{saito2023pic2word} (Fine-tuned) shows only marginal improvement over the pretrained version, suggesting limited benefit from small-scale domain adaptation.

Among ST-CIR methods trained without explicit triplet supervision, VaGFeM~ \cite{le2026figrotd} achieves the best performance in this group, with Recall@1 of 0.0750 and MRR of 0.1142. It consistently outperforms TaskFormer~ \cite{sangkloy2022taskformer} across all metrics, indicating stronger multimodal fusion capability for fine-grained garment attributes.

On the other hand, sketch-only methods perform substantially worse than composite approaches. ZSE-SBIR \cite {lin2023zero} achieves Recall@1 of 0.0285, while S3BIR-DINO~ \cite{s3bir_dino_2025} performs slightly lower at 0.0157. This confirms the importance of guidance on textual attributes for cultural outfit retrieval tasks.

\subsection{Discussion}

% \highlight{chua sua}

The benchmarking results reveal several important insights into the challenges of sketch-text composed cultural outfit retrieval.

\textbf{Impact of Multimodal Composition:}
The large performance gap between SBIR methods and ST-CIR methods demonstrates the importance of textual attribute conditioning. While SBIR methods rely purely on structural sketch similarity, composite models leverage semantic garment attributes such as fabric, decoration, and styling context. This results in substantial gains, with VaGFeM~ \cite{le2026figrotd} achieving nearly $2.6\times$ higher Recall@1 than the best SBIR baseline.

\textbf{Challenges in Zero-Shot and Domain Adaptation:}
Zero-shot methods struggle significantly, with the best model (SEARLE-VIT/L) reaching only 0.0100 Recall@1. This indicates that general vision-language pretraining does not transfer well to the specific intersection of abstract hand-drawn sketches and cultural semantics. Furthermore, the marginal gains seen in Pic2Word (Fine-tuned), even when adapted using sketches from our training set, suggest that lightweight fine-tuning is insufficient to bridge this domain gap. Effective adaptation likely requires larger domain-aligned datasets or more sophisticated self-supervised sketch-image pretraining objectives.

\textbf{Superiority of Fully Supervised CIR Training:}
Supervised CIR models significantly outperform all other paradigms. BLIP4CIR \cite{li2022blip} improves Recall@1 by more than 16\% relative over the best ST-CIR baseline (0.0877 vs 0.0750). The large margin in mAP and MRR also indicates better ranking consistency and attribute grounding capability.

\textbf{Fine-Grained Retrieval Difficulty:}
Even the strongest model achieves Recall@1 below 0.1, while Recall@10 reaches 0.37. This gap highlights the fine-grained difficulty of AoDai retrieval. Many garments share nearly identical silhouettes, differing only in subtle attributes such as embroidery layout, collar design, or fabric texture.

\textbf{Multi-Target Retrieval Complexity:}
The dataset’s multi-target design introduces additional retrieval ambiguity. Because multiple generated images can be valid targets for a single sketch-text query, models must learn attribute-level discrimination rather than instance memorization. The relatively low mAP values across all methods suggest this remains an open challenge.

\textbf{Backbone Architecture Sensitivity:}
The large performance gap between BLIP4CIR~ \cite{li2022blip} and CLIP4CIR~ \cite{baldrati2023composed} suggests that architectures designed for stronger text-visual grounding are better suited to cultural garment retrieval. This is especially important when visual differences are subtle but semantically meaningful.

\textbf{Caption Complexity Trade-off: } 
VietFashion captions are notably detailed, with an average length of 42.46 words. While this ensures high semantic richness and cultural authenticity, the complexity may hinder cross-modal alignment. The superior performance of BLIP4CIR over CLIP4CIR (MRR 0.3950 vs. 0.1908) suggests that architectures with stronger fine-grained text-visual grounding are better at parsing these long-form professional descriptors. Future researches may benefit from evaluating whether condensing these captions improves the focus on core retrieval attributes.

% It is important to note that while the target images in VietFashion are photorealistic, they are synthesized by AI to strictly adhere to the sketch constraints. They are not photographs of physical historical artifacts. This allows for precise control over attributes but introduces a syn-to-real domain gap that users must consider.

\section{Conclusion}

% \highlight{chua sua}

In this work, we presented the VietFashion dataset, a novel benchmark for sketch-text composed image retrieval focused on the Vietnamese traditional garment, Ao Dai. By integrating 650 human sketches with 21,000 synthesized images via a multi-stage pipeline involving Qwen-2.5 and SANA-ControlNet, we established a multi-target environment that effectively addresses false-negative supervision and models realistic retrieval uncertainty. Our benchmarking results reveal a significant performance gap, with BLIP4CIR achieving a state-of-the-art Recall@10 of 0.3703, approximately 9.2 times that of the best zero-shot approach. This disparity highlights the persistent domain gap between abstract sketches and fine-grained cultural imagery in pretrained models, as well as the inherent difficulty of disambiguating visually similar silhouettes based on subtle textual attributes. Moving forward, future research should prioritize self-supervised sketch-image pretraining and more sophisticated fusion mechanisms to enhance fine-grained attribute grounding in specialized cultural domains.

% \section*{Acknowledgments}

\begin{acks}
This research is funded by Vietnam National University - Ho Chi Minh City (VNU-HCM) under Grant Number B2026-18-17.
\end{acks}

% ---------------- BIB ----------------
\bibliographystyle{ACM-Reference-Format}
\balance
\bibliography{refs}

\end{document}